\documentclass[conference]{IEEEtran}
\IEEEoverridecommandlockouts
\usepackage{cite}
\usepackage{amsmath,amssymb,amsfonts}
\usepackage{algorithmic}
\usepackage{graphicx}
\usepackage{textcomp}
\usepackage{xcolor}
\usepackage{tabularray}
\usepackage{url}
\def\BibTeX{{\rm B\kern-.05em{\sc i\kern-.025em b}\kern-.08em
    T\kern-.1667em\lower.7ex\hbox{E}\kern-.125emX}}
\begin{document}

\renewcommand{\thefootnote}{\fnsymbol{footnote}}

\title{Rethinking Gradient-based Adversarial Attacks \\ on Point Cloud Classification}

\author{%
Jun Chen$^{1\ast}$\thanks{$\ast$ Equal Contribution \quad $\dagger$ Corresponding Author} \quad 
Xinke Li$^{2\ast}$ \quad 
Mingyue Xu$^{1}$ \quad 
Chongshou Li$^{1\dagger}$ \quad 
Tianrui Li$^{1}$ \\
$^1$Southwest Jiaotong University $^2$City University of Hong Kong\\
\small
\{2024212410, xumingyue\}@my.swjtu.edu.cn, xinkeli@cityu.edu.hk, \{trli, lics\}@swjtu.edu.cn
}

\maketitle

\begin{abstract}
Gradient-based adversarial attacks are widely used to evaluate the robustness of 3D point cloud classifiers, yet they often rely on uniform update rules that neglect point-wise heterogeneity, leading to perceptible perturbations. We propose two complementary strategies to improve both the effectiveness and imperceptibility of the attack. \textbf{WAAttack} employs weighted gradients to dynamically adjust per-point perturbation magnitudes and uses an adaptive step size strategy to regulate the global perturbation scale. \textbf{SubAttack} partitions the point cloud into subsets and, at each iteration, perturbs only those combinations with high adversarial efficacy and low perceptual saliency. Together, these methods offer a principled refinement of gradient-based attacks for 3D point clouds. Extensive experiments show that our approach consistently outperforms state-of-the-art methods in generating highly imperceptible adversarial examples. The code is available at \url{https://github.com/chenjun0326/WA_SubAttack}.
\end{abstract}


\section{Introduction}
\label{sec:intro}

Despite the significant achievements of Deep Neural Networks (DNNs) in processing 3D multimedia data such as LiDAR point clouds and depth sensors, DNNs have been widely applied in various real-world scenarios including autonomous driving~\cite{zhou2018voxelnet}, indoor navigation~\cite{zhu2017target}, semantic segmentation~\cite{landrieu2018large}, and shape design~\cite{sung2017complementme}. However, these models are particularly vulnerable to adversarial attacks~\cite{xiang2019generating}. Adversarial attacks introduce imperceptible perturbations that cause erroneous output, posing severe risks to safety-critical 3D perception systems. Consequently, developing advanced attack methods is essential for rigorously evaluating the robustness limits of DNNs in practical applications.

\begin{figure}[t]
    \centering
    \includegraphics[width=1\linewidth]{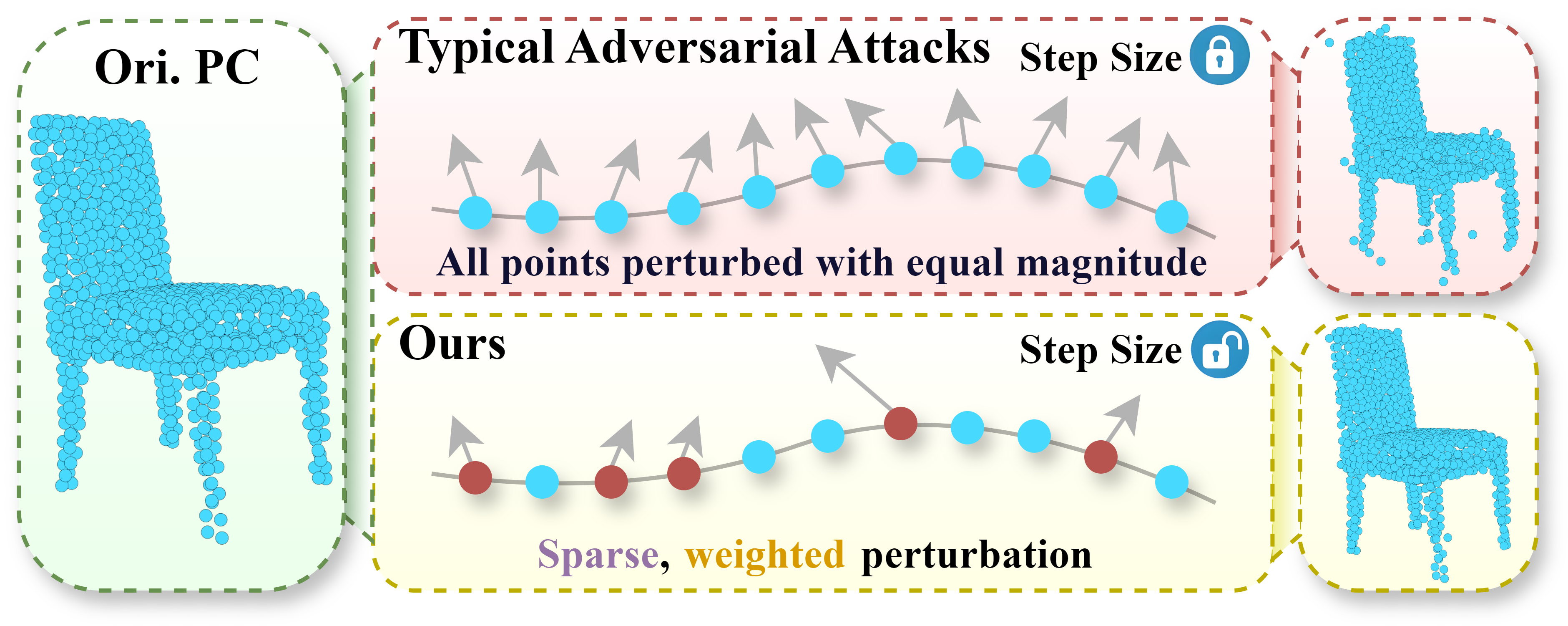}
    \caption{\textbf{Left:} Original point cloud. \textbf{Center:} Typical adversarial attacks perturb all points uniformly (top), while our method applies sparse, weight-guided perturbations with adaptive step sizes (bottom). \textbf{Right:} Resulting adversarial examples under the same budget.}
    \label{fig:fig1_teaser}
    \vspace{-0.25em}
\end{figure}

Although existing methods achieve high success rates~\cite{xiang2019generating}, balancing imperceptibility with attack efficacy remains challenging. Early approaches mitigate geometric distortion via regularization~\cite{wen2020geometry,tsai2020robust} or by using fixed perturbation directions~\cite{huang2022shape,zhang2024curvature}, but often introduce local outliers. Recent works such as SymAttack~\cite{tang2024symattack}, which leverages intrinsic object symmetry, and HiT-Adv~\cite{lou2024hide}, which embeds perturbations in perceptually less sensitive regions, significantly improve visual naturalness. Nevertheless, the prevailing gradient-based attacks still apply a uniform perturbation magnitude across all points and point clouds, as illustrated in Figure~\ref{fig:fig1_teaser}. This ignores both the heterogeneous contribution of individual point gradients and the structural diversity that demands adaptive perturbation intensities, resulting in excessive local distortions and global inconsistency, as shown in the top-right panel of Figure~\ref{fig:fig1_teaser}.

To this end, we propose \textbf{WAAttack}, a novel gradient-based adversarial attack framework through two key mechanisms: (1) a \textit{dynamic weight allocation mechanism} that adaptively modulates the contribution of each point during gradient updates based on its local gradient response, thus mitigating the imbalance of local perturbations caused by uniform update strategies; and (2) an \textit{adaptive step size strategy} that independently adjusts the overall perturbation magnitude for each point cloud sample according to its geometric structure, alleviating inter-sample perturbation distribution disparity. Building on WAAttack, we further introduce \textbf{SubAttack}, as shown in the bottom-center panel of Figure~\ref{fig:fig1_teaser}, which partitions the input point cloud into non-overlapping subsets and, at each iteration, selects the optimal combination of sub-point clouds for perturbation based on a composite score that jointly optimizes attack efficacy and imperceptibility. This avoids redundant modifications to irrelevant regions, thereby enhancing the perceptual imperceptibility of adversarial examples.

Overall, our contributions are threefold:  
\begin{itemize}
    \item We observe that a uniform perturbation fails to account for varying gradient significance across points and diverse structural demands across point clouds.

    \item We propose WAAttack, a gradient-based framework with dynamic weighting and adaptive step sizes, and its enhancement SubAttack that perturbs optimal sub-point cloud combinations.  

    \item Extensive experiments show that our method achieves state-of-the-art imperceptibility while maintaining high attack success.
\end{itemize}

\begin{figure*}[t]
    \centering
    \includegraphics[width=1\linewidth]{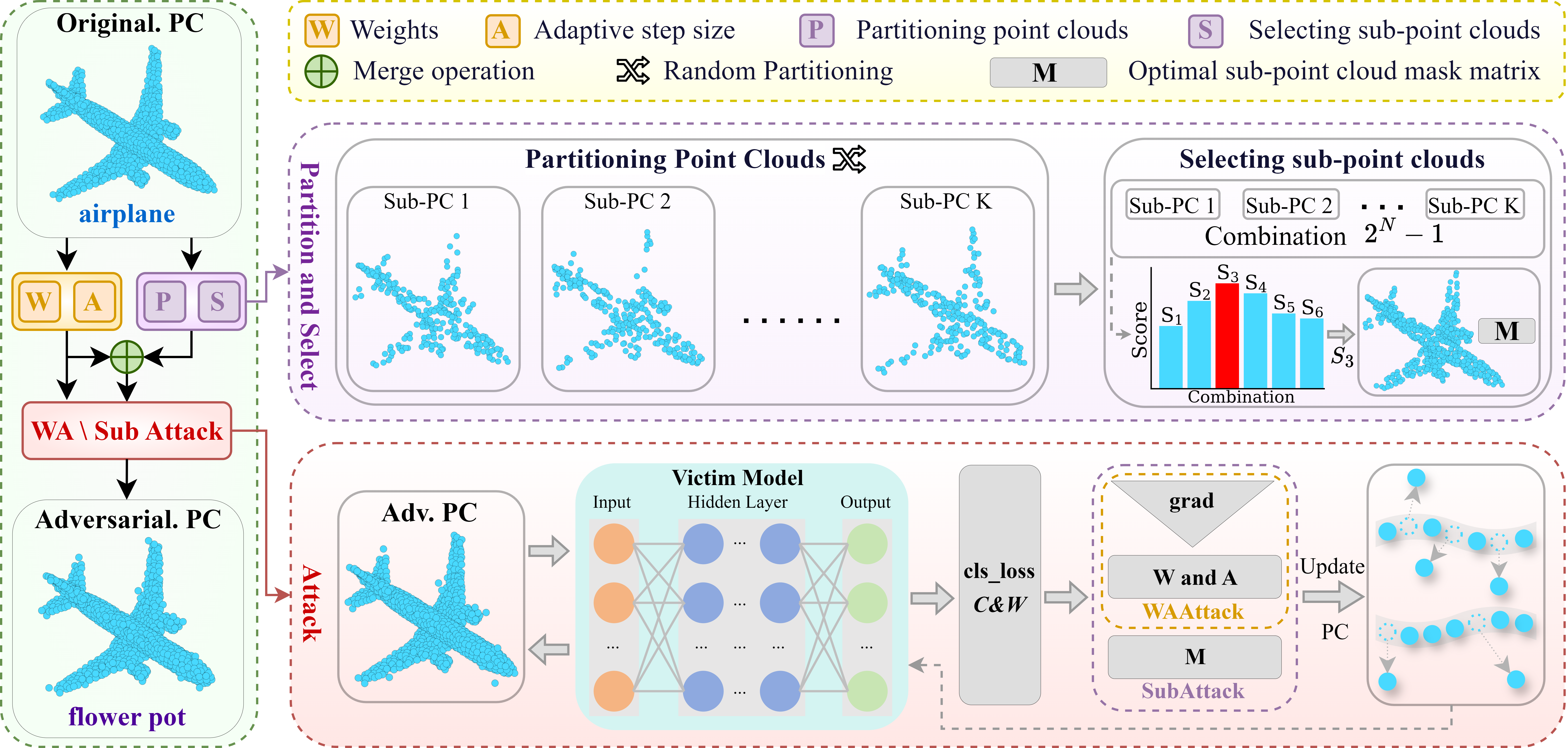}
    \caption{Framework overview of WAAttack and SubAttack. WAAttack enhances adversarial optimization via a weighting mechanism and an adaptive step size. SubAttack randomly partitions the input into sub-point clouds and iteratively applies WAAttack to the most promising one at each step.}
    \label{fig:fig2_framework}
\end{figure*}

\section{Related Work}
\label{Rela}
\textbf{3D Point Cloud Classification.}
Point cloud classification is vital for safety-critical applications. PointNet\cite{qi2017pointnet} pioneers end-to-end learning directly from unordered point sets by using MLPs for per-point feature extraction and max pooling to achieve permutation invariance. DGCNN\cite{wang2019dynamic} significantly advances local structure modeling through dynamic graph construction and iterative feature aggregation via EdgeConv layers. More recently, attention-based architectures, such as PCT\cite{guo2021pct} and geometric-aware models, such as CurveNet\cite{xiang2021walk} further improve performance by capturing long-range dependencies and fine-grained structural priors.

\textbf{Adversarial Attack on Point Cloud Classification.}
Despite their strong performance, these models remain vulnerable to adversarial attacks. Early methods~\cite{xiang2019generating} generate 3D adversarial examples by adding or perturbing points. Subsequent work has mainly aimed at improving the imperceptibility of adversarial perturbations, either through regularization terms~\cite{wen2020geometry}, optimization of perturbation directions~\cite{huang2022shape,zhang2024curvature}, or modeling attacks as geometric deformations of the input shape~\cite{tang2023deep,lou2024hide}. Another research direction focuses on enhancing attack transferability across models~\cite{he2023generating,pang2025towards}, with Pang~et~al.~\cite{pang2025towards} leveraging critical feature guidance to enable effective black-box attacks. In this work, we focus on advancing the imperceptibility of 3D adversarial examples.

\section{Preliminaries}
\label{sec:Prelimi}
\textbf{Typical Gradient-based Attacks.} We consider untargeted attacks on a point cloud classifier $f$, where the attacker has full access to $f$'s architecture and parameters. Given an input point cloud $P \in \mathbb{R}^{N \times 3}$ with true label $y$, where $p_i \in \mathbb{R}^3$ denotes the $i$-th original point and $p'_i$ its perturbed counterpart, the goal is to find $P' = P + \Delta$ such that $f(P') \neq y$ while satisfying $\|\Delta\|_r \leq \epsilon$. This is achieved as follows:
\begin{equation}
    \min_{P'} \mathcal{L}_{\mathrm{cls}}(P') \quad \text{s.t.} \quad \|P' - P\|_r \leq \epsilon.
\end{equation}
Following the C\&W framework~\cite{carlini2017towards}, the loss $\mathcal{L}_{\mathrm{cls}}(P')$ is
\begin{equation}
    \mathcal{L}_{\mathrm{cls}}(P') = \max~\!\bigl(0,\; Z(P')_y - \max_{i \neq y} Z(P')_i + \kappa \bigr),
\end{equation}
where $Z(P')$ denotes the model logits, $y$ the ground-truth label, and $\kappa \geq 0$ a confidence margin. The adversarial point cloud is generated by iteratively updating each point:
\begin{equation}
    p'_i\,^{(t+1)} = p'_i\,^{(t)} - \eta \cdot \vec {\mathbf n}^{(t)},
\end{equation}
where $\vec {\mathbf n}$ is a unit vector encoding the update direction, $\eta$ the step size, and $t$ the iteration index.

\textbf{Discussion.}
Classical gradient-based attacks assume uniform perturbation budgets across point clouds and identical updates for all points. However, perturbation requirements differ across samples due to geometry and model confidence, and within a sample due to varying point importance. Ignoring this heterogeneity yields inefficient attacks with inconsistent success and stealth.

\section{Method}
\label{sec:Method}
To overcome rigid perturbation allocation in classical attacks, we propose WAAttack, which adapts perturbation strength to point cloud sensitivity and per-point importance. Its enhanced variant, SubAttack, further improves performance by selecting the optimal subset combination through a trade-off between loss reduction and geometric distortion. See Figure~\ref{fig:fig2_framework} for an overview.

\subsection{Attack via Weight and Adaptive Step Size (WAAttack)}

\textbf{Point-wise Weighting Mechanism.} To enable fine-grained spatial allocation of perturbations across points, we introduce a gradient-sensitivity-based point-wise weighting mechanism. At iteration $t$, we compute the gradient norm $\|\nabla_{p'^{~(t)}_i} \mathcal{L}_{\mathrm{cls}}\|$ for each point $p'_i$ and scale it by the maximum absolute value of any coordinate gradient over the point cloud:
\begin{equation}
    \omega_i^{(t)} = \frac{\|\nabla_{p'_i\,^{(t)}} \mathcal{L}_{\mathrm{cls}}\|}{\|\nabla_{P'\,^{(t)}} \mathcal{L}_{\mathrm{cls}}\|_{\infty} + \xi},
\end{equation}
where $\xi > 0$ is a small constant to avoid division by zero. The resulting weight reflects the sensitivity of each point to the model’s decision: points in discriminative regions receive stronger perturbations, while less critical points are suppressed, thus improving the attack success rate while maintaining geometric naturalness and stealth.

\textbf{Sample-Aware Step Size Adaptation.}
To accommodate the heterogeneous perturbation requirements across point cloud samples, we propose a dynamic step-size adaptation strategy based on look-ahead loss reduction efficiency. At iteration $t$, we first perform a tentative gradient update to obtain a provisional adversarial sample and compute its classification loss $\mathcal{L}_{\mathrm{cls}}^{(t+1)}$. The relative progress is quantified by:
\begin{equation}
    \rho^{(t)} = \frac{\mathcal{L}_{\mathrm{cls}}^{(t)} - \mathcal{L}_{\mathrm{cls}}^{(t+1)}}{\mathcal{L}_{\mathrm{cls}}^{(0)} + \xi},
\end{equation}
where $\mathcal{L}_{\mathrm{cls}}^{(0)}$ denotes the initial loss. The step size for the next iteration is then adjusted according to:
\begin{equation}
    \eta^{(t+1)} =
    \begin{cases}
        \eta^{(t)}, & \text{if } \rho^{(t)} > \tau, \\
        \alpha \times \, \eta^{(t)}, & \text{if } 0 < \rho^{(t)} \leq \tau, \\
        \beta \times \, \eta^{(t)}, & otherwise,
    \end{cases}
\end{equation}
where $\tau = c / T_{\max}$, $T_{\max}$ is the maximum number of iterations, $c > 0$ is a sensitivity coefficient, and $\alpha > 1$, $\beta \in (0,1)$ are step-size amplification and attenuation factors, respectively. This mechanism enables the attack to automatically balance convergence speed and perturbation stability based on per-sample optimization dynamics.

In summary, the adversarial point update in our WAAttack framework is given by:
\begin{equation}
    p_i^{\prime\,(t+1)} = p_i^{\prime\,(t)} - \eta^{(t)} \cdot \, \omega_i^{(t)} \cdot \, \vec {\mathbf n}^{(t)},
\end{equation}
which jointly incorporates per-point sensitivity through $\omega_i^{(t)}$ and adaptive step size control via $\eta^{(t)}$.

\subsection{Partitioning for Enhanced Attack Performance (SubAttack)}
\textbf{Point Cloud Partitioning.} Recent work~\cite{shi2022shape} shows that sparse perturbations, which restrict adversarial noise to critical points, improve attack effectiveness while reducing perceptual and geometric distortion. Inspired by this, we achieve implicit sparsity by perturbing only the most vulnerable point subsets rather than the entire point cloud. Specifically, we randomly partition $P \in \mathbb{R}^{N \times 3}$ into $K$ disjoint and collectively exhaustive subsets: for each sample, a random permutation of point indices is generated, with the first $N \bmod K$ subsets receiving $\lfloor N/K \rfloor + 1$ points and the rest $\lfloor N/K \rfloor$, ensuring near-uniform sizes. The partitioning is re-sampled independently per attack to enhance diversity and avoid structural bias.

\textbf{Optimal Subset Selection.}
To balance attack effectiveness and perturbation stealth, we select the subset combination that maximizes a composite score $S$ over all non-empty combinations of the $K$ partitioned subsets. For each candidate combination, we perform a single-step gradient update only on its covered points and compute:
\begin{equation}
    S = \Delta \mathcal{L}_{\mathrm{cls}} - \lambda D,
\end{equation}
where $\Delta \mathcal{L}_{\mathrm{cls}} = \mathcal{L}_{\mathrm{cls}}^{\text{orig}} - \mathcal{L}_{\mathrm{cls}}^{\text{new}}$ quantifies the reduction in classification loss (indicating attack potential), and $D$ measures geometric distortion:
\begin{equation}
    D = w_c D_c + w_h D_h + w_l D_l.
\end{equation}
Here, $D_c$, $D_h$, and $D_l$ denote the Chamfer distance~\cite{fan2017point}, Hausdorff distance~\cite{taha2015metrics}, and overall $l_2$ perturbation norm, respectively, with positive weights $w_c, w_h, w_l > 0$. The scalar $\lambda > 0$ controls the trade-off between attack strength and perceptual fidelity.

Let $\mathcal{C}^{*(t)}$ denote the optimal subset combination at iteration $t$, which achieves the highest score $S^{(t)}$. We construct a binary mask $\mathbf{M}^{(t)} \in \{0,1\}^{N}$ such that $M_i^{(t)} = 1$ if point $p_i$ belongs to any subset in $\mathcal{C}^{*(t)}$, and $M_i^{(t)} = 0$ otherwise. This mask explicitly identifies the most vulnerable regions for the targeted perturbation. SubAttack then updates only these selected points as:
\begin{equation}
    p_i^{\prime\,(t+1)} = p_i^{\prime\,(t)} - \eta^{(t)} \cdot \, \omega_i^{(t)} \cdot \, \vec {\mathbf{n}}^{(t)} \cdot \, M_i^{(t)},
\end{equation}
ensuring sparse, effective perturbations with controlled geometric distortion.

\begin{table*}
\centering
\caption{Attack performance in terms of attack success rate (ASR\,$\uparrow$), geometric distortion ($D_c$\,$\downarrow$, $D_h$\,$\downarrow$, $D_l$\,$\downarrow$), and efficiency (A.T\,$\downarrow$) on ModelNet40 (MN) and ScanObjectNN (SONN). Best results are in \textbf{bold}, second-best are \underline{underlined}.}
\begin{tblr}{
  cells = {c},
  colsep=0.08em,
  cell{1}{1} = {r=2}{},
  cell{1}{2} = {r=2}{},
  cell{1}{3} = {c=5}{},
  cell{1}{8} = {c=5}{},
  cell{1}{13} = {c=5}{},
  cell{1}{18} = {c=5}{},
  cell{3}{1} = {r=10}{},
  cell{13}{1} = {r=10}{},
  vline{2,3,8,13,18} = {1-22}{},
  hline{2} = {3-22}{},
  hline{1,3,23} = {-}{},
  hline{11,21} = {2-22}{},
  hline{13} = {-}{},
}
Dataset      & {Attack\\Method}   & PointNet       &                                            &                                            &                 &               & DGCNN~         &                                            &                                            &                 &               & PCT            &                                            &                                            &                 &               & CurveNet       &                                            &                                            &                 &               \\
             &                    & {ASR\\(\%)}    & {$D_c$\\({10\textsuperscript{-4}})} & {$D_h$\\({10\textsuperscript{-2}})} & $D_l$           & {A.T\\(s)}    & {ASR\\(\%)}    & {$D_c$\\({10\textsuperscript{-4}})} & {$D_h$\\({10\textsuperscript{-2}})} & $D_l$           & {A.T\\(s)}    & {ASR\\(\%)}    & {$D_c$\\({10\textsuperscript{-4}})} & {$D_h$\\({10\textsuperscript{-2}})} & $D_l$           & {A.T\\(s)}    & {ASR\\(\%)}    & {$D_c$\\({10\textsuperscript{-4}})} & {$D_h$\\({10\textsuperscript{-2}})} & $D_l$           & {A.T\\(s)}    \\
MN   & 3D-Adv~\cite{xiang2019generating}            & \textbf{100.0} & 4.60                                       & 3.75                                       & 0.7680          & 93.62         & \textbf{100.0} & 22.96                                      & 2.87                                       & 1.3881          & 74.66         & \textbf{100.0} & 14.35                                      & 2.98                                       & 1.3664          & 1729.05       & \textbf{100.0} & 15.79                                      & 4.42                                       & 1.2776          & 1350.87       \\
             & JGBA~\cite{wen2020geometry}              & \underline{99.9}           & 27.68                                      & 6.74                                       & 1.5250          & 5.06          & \textbf{100.0} & 58.99                                      & 6.03                                       & 2.8783          & 5.20          & \textbf{100.0} & 58.23                                      & 5.00                                       & 2.7697          & 20.83         & \textbf{100.0} & 61.59                                      & 6.04                                       & 2.8550          & 26.99         \\
             & GeoA\textsuperscript{3}~\cite{wen2020geometry}              & \textbf{100.0} & 6.55                                       & \textbf{0.56}                              & 1.3586          & 97.24         & \textbf{100.0} & 10.60                                      & 0.59                                       & 2.1977          & 83.65         & \textbf{100.0} & 8.76                                       & 0.89                                       & 2.1009          & 1767.70       & \textbf{100.0} & 8.64                                       & 0.94                                       & 2.2319          & 1193.27       \\
             & GSDA~\cite{hu2022exploring}               & \textbf{100.0} & 5.04                                       & 1.31                                       & 1.3167          & 99.70         & \textbf{100.0} & 7.86                                       & 0.75                                       & 1.9348          & 102.14        & \textbf{100.0} & 6.58                                       & 1.04                                       & 2.1221          & 1789.37       & \textbf{100.0} & 5.90                                       & 0.98                                       & 2.2954          & 1322.39       \\
             & SI-Adv~\cite{huang2022shape}             & \textbf{100.0} & 3.72                                       & 2.57                                       & 0.5607          & 0.99          & \textbf{100.0} & 14.91                                      & 1.96                                       & 1.0963          & \underline{0.75}          & \textbf{100.0} & 10.77                                      & 2.49                                       & 0.8682          & 10.14         & \underline{99.8}           & 12.30                                      & 3.32                                       & 0.9997          & \underline{9.78}          \\
             & PF-Attack~\cite{he2023generating}          & 95.5           & 21.24                                      & 2.44                                       & 1.7487          & 10.55         & \underline{99.7}           & 50.43                                      & 2.50                                       & 2.8386          & 9.93          & 99.4           & 45.57                                      & 2.51                                       & 2.7021          & 179.86        & 93.2           & 44.36                                      & 2.38                                       & 2.3906          & 149.90        \\
             & CIM~\cite{zhang2024curvature}                & \underline{99.9}           & 3.81                                       & 2.51                                       & 0.5600          & 1.36          & \underline{99.7}           & 14.89                                      & 2.21                                       & 1.0929          & 1.19          & \underline{99.8}           & 10.62                                      & 2.62                                       & 0.8591          & \underline{9.88}          & 99.7           & 12.01                                      & 3.31                                       & 0.9764          & 10.05         \\
             & HiT-Adv~\cite{lou2024hide}            & \textbf{100.0} & 44.67                                      & 3.97                                       & 1.4223          & 35.37         & \textbf{100.0} & 59.78                                      & 5.85                                       & 17054           & 43.01         & \textbf{100.0} & 72.10                                      & 5.18                                       & 1.7231          & 215.66        & \textbf{100.0} & 39.55                                      & 7.39                                       & 1.2153          & 155.72        \\
             & \textbf{WAAttack}  & \textbf{100.0} & \underline{0.90}                                       & 0.78                                       & \underline{0.1812}          & \textbf{0.26} & \underline{99.7}           & \underline{3.20}                                       &\underline{0.55}                                       & \underline{0.3321}          & \textbf{0.18} & \underline{99.8}           & \underline{2.59}                                       & \underline{0.80}                                       & \underline{0.2977}          & \textbf{4.94} & 98.5           & \underline{2.23}                                       & \underline{0.91}                                       & \underline{0.2727}          & \textbf{4.96} \\
             & \textbf{SubAttack} & \textbf{100.0} & \textbf{0.81}                              & \underline{0.72}                                       & \textbf{0.1632} & \underline{0.98}          & \textbf{100.0} & \textbf{2.89}                              & \textbf{0.51}                              & \textbf{0.2958} & 0.98          & \textbf{100.0} & \textbf{2.39}                              & \textbf{0.77}                              & \textbf{0.2749} & 31.34         & \underline{99.8}           & \textbf{2.20}                              & \textbf{0.85}                              & \textbf{0.2612} & 23.96         \\
SONN & 3D-Adv~\cite{xiang2019generating}             & \textbf{100.0} & 6.38                                       & 3.95                                       & 0.7346          & 88.16         & \textbf{100.0} & 15.88                                      & 2.53                                       & 1.1265          & 73.13         & \textbf{100.0} & 13.26                                      & 2.79                                       & 1.2437          & 1717.86       & \textbf{100.0} & 19.36                                      & 4.12                                       & 1.5364          & 1306.28       \\
             & JGBA~\cite{ma2020efficient}               & 96.9           & 53.70                                      & 8.14                                       & 2.3090          & 5.19          & \textbf{100.0} & 49.25                                      & 6.23                                       & 2.7712          & 5.25          & \textbf{100.0} & 46.43                                      & 5.24                                       & 2.7858          & 20.93         & \textbf{100.0} & 59.79                                      & 6.63                                       & 3.2010          & 27.30         \\
             & GeoA\textsuperscript{3}~\cite{wen2020geometry}              & \textbf{100.0} & 7.42                                       & \textbf{0.75}                              & 1.6682          & 91.32         & \textbf{100.0} & 5.45                                       & 0.28                                       & 1.8935          & 98.68         & \textbf{100.0} & 6.17                                       & \textbf{0.34}                              & 1.6391          & 1747.91       & \textbf{100.0} & 7.70                                       & \textbf{0.58}                              & 2.0549          & 1336.25       \\
             & GSDA~\cite{hu2022exploring}               & \textbf{100.0} & 5.80                                       & 1.34                                       & 1.8148          & 108.12        & \textbf{100.0} & 4.80                                       & 0.46                                       & 2.0109          & 93.42         & \textbf{100.0} & 4.78                                       & 0.60                                       & 2.1551          & 1619.43       & \textbf{100.0} & 7.06                                       & 1.05                                       & 2.1978          & 1236.19       \\
             & SI-Adv~\cite{huang2022shape}             & 97.9           & 3.99                                       & 3.21                                       & 0.5726          & \underline{0.90}          & \underline{99.8}           & 8.42                                       & 1.71                                       & 0.9178          & \underline{0.78}          & \textbf{100.0} & 7.52                                       & 2.18                                       & 0.8552          & 9.97          & \underline{99.1}           & 10.76                                      & 3.45                                       & 1.0198          & 9.90          \\
             & PF-Attack~\cite{he2023generating}          & 88.3           & 29.71                                      & 2.32                                       & 1.8787          & 11.58         & 98.5           & 50.34                                      & 2.31                                       & 2.4197          & 10.16         & 97.8           & 44.71                                      & 2.38                                       & 2.3356          & 183.24        & 92.9           & 42.35                                      & 2.39                                       & 2.2965          & 148.22        \\
             & CIM~\cite{zhang2024curvature}                & 97.2           & 3.70                                       & 2.85                                       & 0.5285          & 1.40          & \underline{99.8}           & 7.78                                       & 1.72                                       & 0.8298          & 1.23          & 99.5           & 7.25                                       & 2.17                                       & 0.7992          & \underline{9.61}          & 99.0           & 10.07                                      & 3.16                                       & 0.9246          & \underline{9.25}          \\
             & HiT-Adv~\cite{lou2024hide}            & 96.4           & 55.72                                      & 3.58                                       & 1.5179          & 44.40         & 98.7           & 19.22                                      & 2.87                                       & 1.0136          & 51.38         & 97.9           & 43.10                                      & 3.21                                       & 1.3584          & 185.98        & 96.8           & 46.28                                      & 4.79                                       & 1.2655          & 166.37        \\
             & \textbf{WAAttack}  & 99.3           & \underline{1.65}                                       & 1.16                                       & \underline{0.2237}          & \textbf{0.20} & 99.7           & \underline{0.91}                                       & \underline{0.18}                                       & \underline{0.1761}          & \textbf{0.18} & \underline{99.8}           & \underline{1.64}                                       & 0.48                                       & \underline{0.2397}\underline{    }      & \textbf{4.41} & 98.0           & \underline{2.47}                                       & 1.16                                       & \underline{0.3129}          & \textbf{4.84} \\
             & \textbf{SubAttack} & \underline{99.5}           & \textbf{1.60}                              & \underline{1.08}                                       & \textbf{0.2151} & \underline{0.90}          & \textbf{100.0} & \textbf{0.71}                              & \textbf{0.14}                              & \textbf{0.1369} & \underline{0.78}          & \textbf{100.0} & \textbf{1.43}                              & \underline{0.45}                                       & \textbf{0.2115} & 26.26         & \textbf{100.0} & \textbf{2.27}                              & \underline{1.00}                                       & \textbf{0.2856} & 21.16         
\end{tblr}
\label{tab:table1}
\end{table*}

\section{Experimental Results}
\label{sec:Exper}
\subsection{Experimental Setup}
\textbf{Implementation Details.} We adopt the $\ell_\infty$ perturbation norm with a budget of $\epsilon = 0.16$, maximum iterations $T_{\max} = 50$, initial step size $\eta = 0.007$, and constant $\xi = 10^{-8}$. For WAAttack, the adaptive step size parameters are set to sensitivity coefficient $c = 2$, amplification factor $\alpha = 1.6$, and attenuation factor $\beta = 0.8$. SubAttack uses $K = 4$ subsets, distortion weights $w_c = 1000$, $w_h = 100$, $w_\ell = 0.1$, and trade-off coefficient $\lambda = 0.1$. All experiments are conducted on a server equipped with four NVIDIA A100 GPUs.


\textbf{Datasets.} We evaluate our method on two widely used 3D multimedia benchmark datasets.  ModelNet40~\cite{wu20153d} consists of 12,311 synthetic CAD models across 40 common object categories and serves as a standard benchmark for 3D shape understanding. ScanObjectNN~\cite{uy2019revisiting} comprises 15,000 real-world scanned indoor object point clouds with occlusions and background clutter, making it more representative of practical multimedia scenarios. All point clouds are uniformly sampled to 1,024 points and normalized into a unit cube.

\textbf{Victim Models and Baselines.}
We evaluate our attack on four widely adopted point cloud classifiers: PointNet~\cite{qi2017pointnet}, PointNet++~\cite{qi2017pointnet++}, DGCNN~\cite{wang2019dynamic}, and CurveNet~\cite{xiang2021walk}. We compare against eight baseline attacks: seven point-based perturbation methods (3D-Adv~\cite{xiang2019generating}, JGBA~\cite{ma2020efficient}, GeoA\textsuperscript{3}~\cite{wen2020geometry}, GSDA~\cite{hu2022exploring}, SI-Adv~\cite{huang2022shape}, PF-Attack~\cite{he2023generating}, CIM~\cite{zhang2024curvature}) and one deformation-based attack, HiT-Adv~\cite{lou2024hide}.

\textbf{Evaluation Metrics.} We evaluate attacks using five metrics: Attack Success Rate (ASR) for effectiveness; Chamfer distance~\cite{fan2017point} and Hausdorff distance~\cite{taha2015metrics} for geometric fidelity; $\ell_2$ norm for average perturbation magnitude; and average time cost (A.T) for computational efficiency.

\subsection{Comprehensive Attack Evaluation}
\textbf{Effectiveness and Imperceptibility.} As shown in Table~\ref{tab:table1}, our methods achieve the best or second-best performance on nearly all metrics on both ModelNet40 and ScanObjectNN, significantly outperforming state-of-the-art approaches such as CIM~\cite{zhang2024curvature} and HiT-Adv~\cite{lou2024hide}. Specifically, \textbf{WAAttack} maintains high ASR while substantially reducing $D_c$, $D_h$, and $D_l$, and attains the lowest average runtime (A.T) among all methods. Building upon WAAttack, \textbf{SubAttack} further improves both ASR and imperceptibility through sub-point cloud partitioning, with A.T. remaining at a practical level and clearly better than most baselines. These results demonstrate the exceptional balance of attack effectiveness, stealth, and computational efficiency achieved by our framework.

\textbf{Visualization Analysis.} To better illustrate the imperceptibility of our adversarial point clouds, we visualize the outputs of different attack methods on the same input, as shown in Figure~\ref{fig:visualization_dgcnn}. Compared to baselines, our approach generates adversarial samples with fewer outliers and a more uniform distribution. This is enabled by the weight term, adaptive step size, and sub-point cloud optimization, which together improve imperceptibility without causing noticeable structural distortions. In particular, although HiT-Adv is a deformation-based attack and also exhibits few outliers, its geometric deformations remain visually perceptible.

\begin{figure}[t]
    \centering
    \includegraphics[width=1\linewidth]{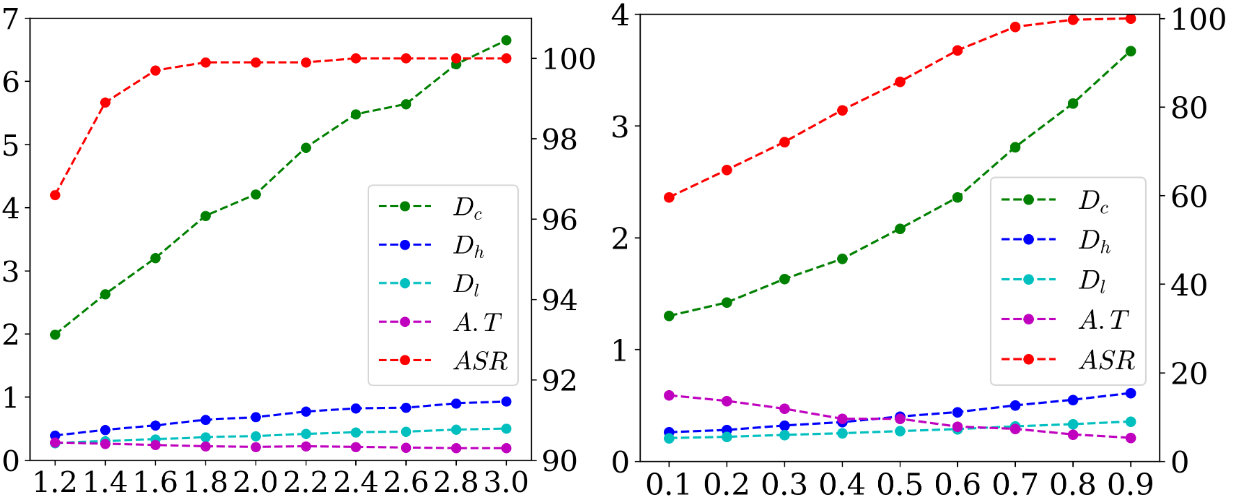}
    \caption{Ablation study of $\alpha$ (left) and $\beta$ (right) on DGCNN. Left y-axis: $D_c~(10^{-4})$, $D_h~(10^{-2})$, $D_l$, $A.T~(s)$; right y-axis: ASR (\%).}
    \label{fig:fig4_alpha_beta}
\end{figure}

\begin{figure*}[t]
    \centering
    \includegraphics[width=1\linewidth]{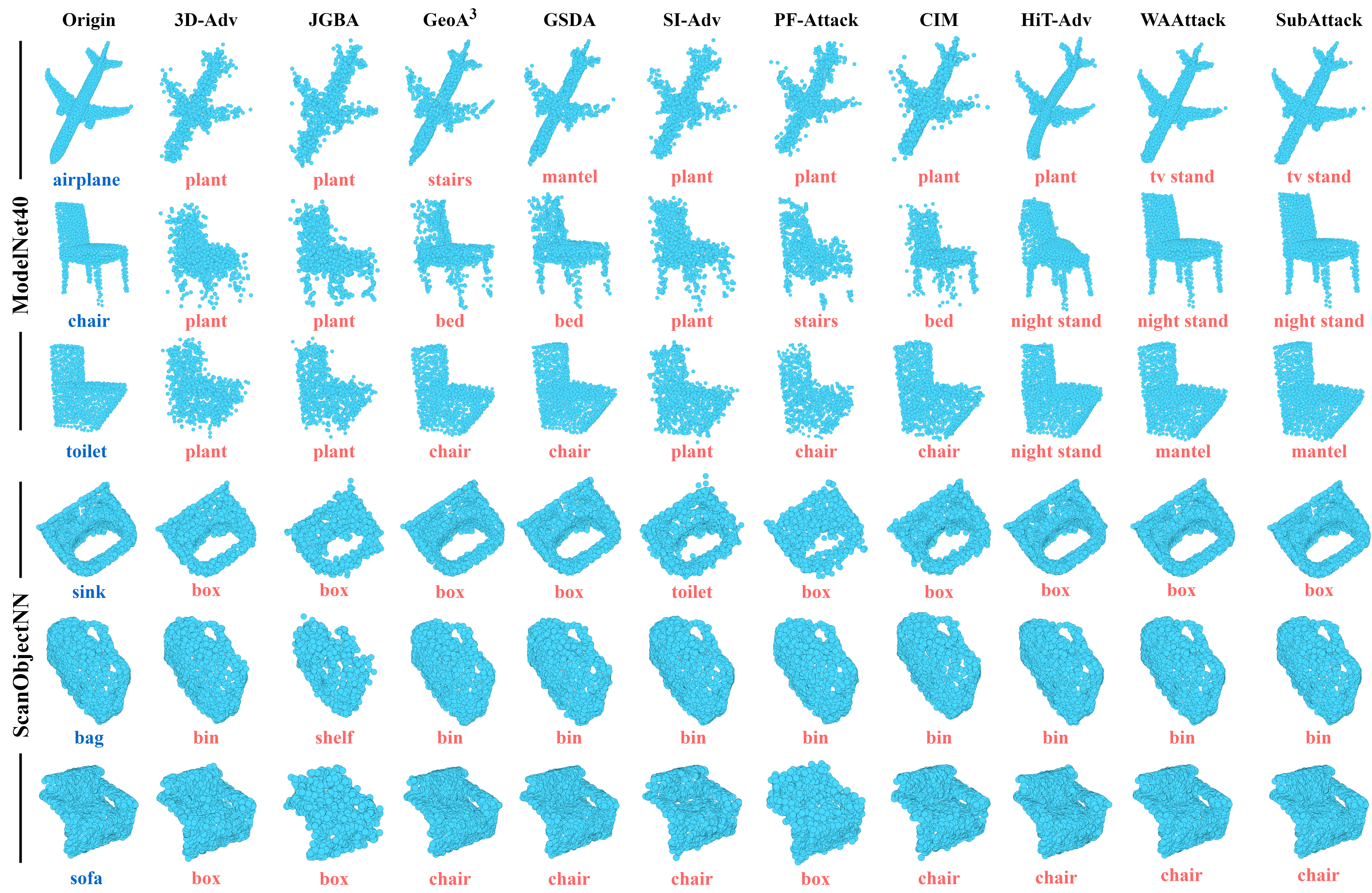}
    \vspace{-2em}
    \caption{Visualization of adversarial point clouds generated by different attack methods on ModelNet40 and ScanObjectNN when targeting DGCNN.}
    \label{fig:visualization_dgcnn}
    \vspace{-1em}
\end{figure*}

\begin{figure}[t]
    \centering
    \includegraphics[width=1\linewidth]{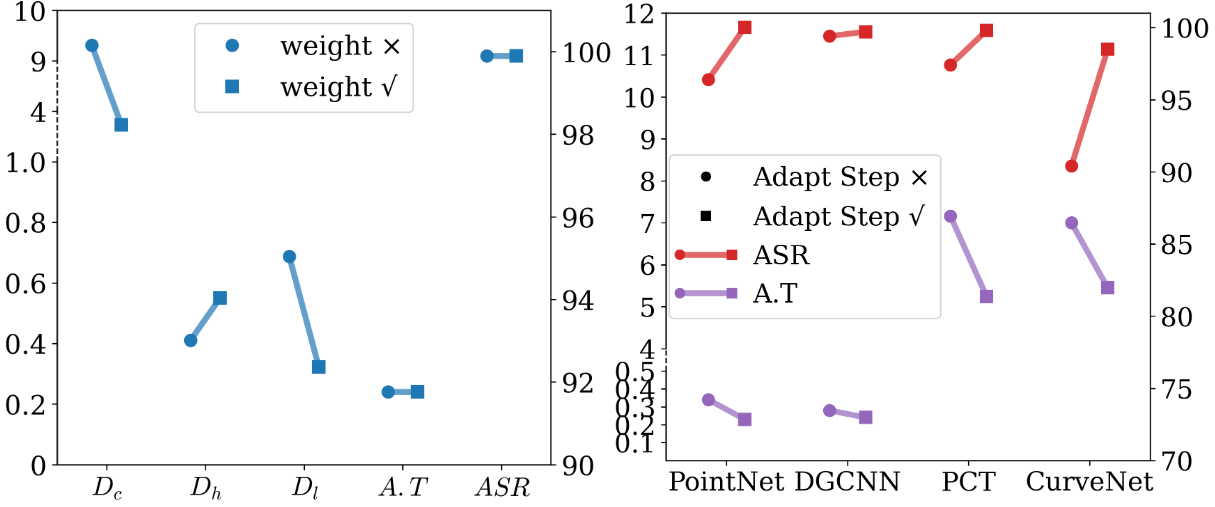}
    \vspace{-2em}
    \caption{Effectiveness of the weight term (left) and adaptive step size (right). Left y-axis: $D_c~(10^{-4})$, $D_h~(10^{-2})$, $D_l$, $A.T~(s)$; right y-axis: ASR (\%).}
    \vspace{-1em}
    \label{fig:fig5_weight_step}
\end{figure}

\subsection{Ablation studies on ModelNet40}
\textbf{Ablation on $\alpha$ and $\beta$.} Based on Figure~\ref{fig:fig4_alpha_beta}, we set $\alpha = 1.6$ and $\beta = 0.8$. At $\alpha = 1.6$, ASR saturates with low distortion; larger values only increase perturbation. At $\beta = 0.8$, ASR is maximized without compromising imperceptibility, yielding the best balance of effectiveness and stealth.

\begin{table}
\centering
\caption{Ablation on $K$ with DGCNN.}
\begin{tabular}{cccccc} 
\hline
K                                                                     & \begin{tabular}[c]{@{}c@{}}ASR\\(\%)\end{tabular}                                               & \begin{tabular}[c]{@{}c@{}}$D_c$\\(10\textsuperscript{-4})\end{tabular}                         & \begin{tabular}[c]{@{}c@{}}$D_h$\\(10\textsuperscript{-2})\end{tabular}                        & $D_l$                                                                                                           & \begin{tabular}[c]{@{}c@{}}A.T\\(s)\end{tabular}                                                  \\ 
\hline
\begin{tabular}[c]{@{}c@{}}1\\2\\3\\\textbf{4}\\5\\6\\7\\8\end{tabular} & \begin{tabular}[c]{@{}c@{}}99.7\\99.4\\99.3\\\textbf{99.4}\\99.3\\99.4\\99.1\\99.6\end{tabular} & \begin{tabular}[c]{@{}c@{}}3.20\\2.77\\2.46\\\textbf{2.10}\\1.99\\1.86\\1.79\\1.69\end{tabular} & \begin{tabular}[c]{@{}c@{}}0.55\\0.50\\0.46\\\textbf{0.40}\\0.39\\0.37\\0.35\\0.34\end{tabular} & \begin{tabular}[c]{@{}c@{}}0.3321\\0.3059\\0.2872\\\textbf{0.2671}\\0.2584\\0.2505\\0.2446\\0.2368\end{tabular} & \begin{tabular}[c]{@{}c@{}}0.24\\1.36\\1.60\\\textbf{2.01}\\2.82\\4.47\\7.69\\13.79\end{tabular}  \\
\hline
\end{tabular}
\label{tab:table2}
\end{table}

\begin{figure}[htbp]
    \centering
    \includegraphics[width=0.9\linewidth]{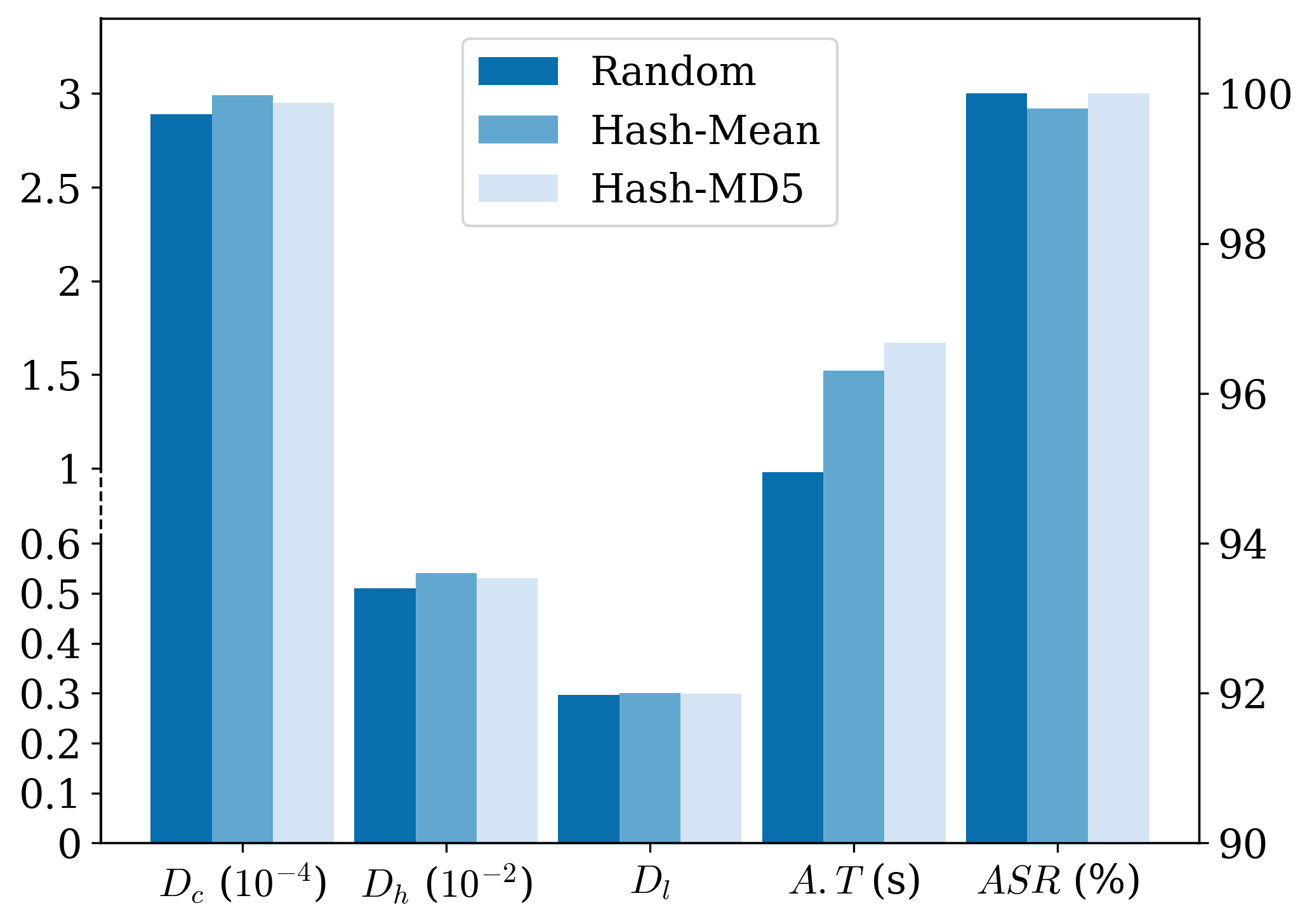}
    \vspace{-1em}
    \caption{Investigation on Different Division Methods. Victim model: DGCNN. Left y-axis: $D_c~(10^{-4})$, $D_h~(10^{-2})$, $D_l$, $A.T~(s)$; right y-axis: ASR (\%).}
    \vspace{-1em}
    \label{fig:fig6_partition}
\end{figure}

\textbf{Core Components.}  
We evaluate the weight term and adaptive step size via ablation. As shown in Figure~\ref{fig:fig5_weight_step} left, on DGCNN the weight term preserves ASR and A.T, slightly increases $D_h$, but greatly reduces $D_c$ and $D_l$, confirming its effectiveness in perturbation suppression. As shown in Figure~\ref{fig:fig5_weight_step} right, the adaptive step size increases ASR and lowers A.T across PointNet, PointNet++, DGCNN, and CurveNet, demonstrating improved attack strength and efficiency.

\textbf{Effect of $K$. }As shown in Table~\ref{tab:table2}, $K = 4$ maintains ASR, significantly improves imperceptibility, and avoids the marginal gains and high overhead of larger $K$, making it the optimal choice.

\textbf{Comparison of Sub-point Cloud Partitioning Strategies.}  As shown in Figure~\ref{fig:fig6_partition}, random partitioning outperforms hash-based methods~\cite{zhang2023pointcert} in terms of ASR, imperceptibility, and efficiency. This is because random partitioning independently re-samples sub-point clouds for each attack, introducing beneficial randomness that enhances perturbation diversity and avoids structural bias caused by fixed grouping. More experiments (e.g., robustness, ablation study on parameters $c$, $\lambda$, and weight term design) are provided in supplementary materials.

\section{Conclusion} \label{sec:Comc}
In this paper, we propose an efficient attack methd for 3D point clouds, integrating gradient weighting, adaptive step size, and random sub-point cloud partitioning. 
Experiments on two datasets show it outperforms state-of-the-art approaches, striking a strong balance among effectiveness, imperceptibility, and practicality. 

\section*{Acknowledgements} 
This project is partially supported by Sichuan Science and Technology Program (Grant No: 2026YFHZ0077). 


\bibliographystyle{IEEEbib}
\bibliography{references}

@inproceedings{xiang2019generating,
  title={Generating 3d adversarial point clouds},
  author={Xiang, Chong and Qi, Charles R and Li, Bo},
  booktitle={Proceedings of the IEEE/CVF Conference on Computer Vision and Pattern Recognition},
  pages={9136--9144},
  year={2019}
}

@inproceedings{tsai2020robust,
  title={Robust adversarial objects against deep learning models},
  author={Tsai, Tzungyu and Yang, Kaichen and Ho, Tsung-Yi and Jin, Yier},
  booktitle={Proceedings of the AAAI Conference on Artificial Intelligence},
  volume={34},
  pages={954--962},
  year={2020}
}

@inproceedings{ma2020efficient,
  title={Efficient joint gradient based attack against sor defense for 3d point cloud classification},
  author={Ma, Chengcheng and Meng, Weiliang and Wu, Baoyuan and Xu, Shibiao and Zhang, Xiaopeng},
  booktitle={Proceedings of the 28th ACM International Conference on Multimedia},
  pages={1819--1827},
  year={2020}
}

@article{wen2020geometry,
  title={Geometry-aware generation of adversarial point clouds},
  author={Wen, Yuxin and Lin, Jiehong and Chen, Ke and Chen, CL Philip and Jia, Kui},
  journal={IEEE Transactions on Pattern Analysis and Machine Intelligence},
  volume={44},
  pages={2984--2999},
  year={2020},
  publisher={IEEE}
}

@inproceedings{hu2022exploring,
  title={Exploring the devil in graph spectral domain for 3d point cloud attacks},
  author={Hu, Qianjiang and Liu, Daizong and Hu, Wei},
  booktitle={European Conference on Computer Vision},
  pages={229--248},
  year={2022},
  organization={Springer}
}

@inproceedings{huang2022shape,
  title={Shape-invariant 3D adversarial point clouds},
  author={Huang, Qidong and Dong, Xiaoyi and Chen, Dongdong and Zhou, Hang and Zhang, Weiming and Yu, Nenghai},
  booktitle={Proceedings of the IEEE/CVF Conference on Computer Vision and Pattern Recognition},
  pages={15335--15344},
  year={2022}
}

@inproceedings{he2023generating,
  title={Generating transferable 3d adversarial point cloud via random perturbation factorization},
  author={He, Bangyan and Liu, Jian and Li, Yiming and Liang, Siyuan and Li, Jingzhi and Jia, Xiaojun and Cao, Xiaochun},
  booktitle={Proceedings of the AAAI Conference on Artificial Intelligence},
  volume={37},
  pages={764--772},
  year={2023}
}

@inproceedings{zhang2024curvature,
  title={Curvature-invariant adversarial attacks for 3d point clouds},
  author={Zhang, Jianping and Gu, Wenwei and Huang, Yizhan and Jiang, Zhihan and Wu, Weibin and Lyu, Michael R},
  booktitle={Proceedings of the AAAI Conference on Artificial Intelligence},
  volume={38},
  pages={7142--7150},
  year={2024}
}

@inproceedings{lou2024hide,
  title={Hide in thicket: Generating imperceptible and rational adversarial perturbations on 3d point clouds},
  author={Lou, Tianrui and Jia, Xiaojun and Gu, Jindong and Liu, Li and Liang, Siyuan and He, Bangyan and Cao, Xiaochun},
  booktitle={Proceedings of the IEEE/CVF Conference on Computer Vision and Pattern Recognition},
  pages={24326--24335},
  year={2024}
}

@inproceedings{zhang2023pointcert,
  title={Pointcert: Point cloud classification with deterministic certified robustness guarantees},
  author={Zhang, Jinghuai and Jia, Jinyuan and Liu, Hongbin and Gong, Neil Zhenqiang},
  booktitle={Proceedings of the IEEE/CVF Conference on Computer Vision and Pattern Recognition},
  pages={9496--9505},
  year={2023}
}

@inproceedings{wu20153d,
  title={3d shapenets: A deep representation for volumetric shapes},
  author={Wu, Zhirong and Song, Shuran and Khosla, Aditya and Yu, Fisher and Zhang, Linguang and Tang, Xiaoou and Xiao, Jianxiong},
  booktitle={Proceedings of the IEEE Conference on Computer Vision and Pattern Recognition},
  pages={1912--1920},
  year={2015}
}

@inproceedings{qi2017pointnet,
  title={Pointnet: Deep learning on point sets for 3d classification and segmentation},
  author={Qi, Charles R and Su, Hao and Mo, Kaichun and Guibas, Leonidas J},
  booktitle={Proceedings of the IEEE Conference on Computer Vision and Pattern Recognition},
  pages={652--660},
  year={2017}
}

@article{qi2017pointnet++,
  title={Pointnet++: Deep hierarchical feature learning on point sets in a metric space},
  author={Qi, Charles Ruizhongtai and Yi, Li and Su, Hao and Guibas, Leonidas J},
  journal={Advances in Neural Information Processing Systems},
  volume={30},
  year={2017}
}

@article{wang2019dynamic,
  title={Dynamic graph cnn for learning on point clouds},
  author={Wang, Yue and Sun, Yongbin and Liu, Ziwei and Sarma, Sanjay E and Bronstein, Michael M and Solomon, Justin M},
  journal={ACM Transactions on Graphics},
  volume={38},
  pages={1--12},
  year={2019},
  publisher={Acm New York, NY, USA}
}

@inproceedings{xiang2021walk,
  title={Walk in the cloud: Learning curves for point clouds shape analysis},
  author={Xiang, Tiange and Zhang, Chaoyi and Song, Yang and Yu, Jianhui and Cai, Weidong},
  booktitle={Proceedings of the IEEE/CVF International Conference on Computer Vision},
  pages={915--924},
  year={2021}
}

@inproceedings{fan2017point,
  title={A point set generation network for 3d object reconstruction from a single image},
  author={Fan, Haoqiang and Su, Hao and Guibas, Leonidas J},
  booktitle={Proceedings of the IEEE Conference on Computer Vision and Pattern Recognition},
  pages={605--613},
  year={2017}
}

@article{taha2015metrics,
  title={Metrics for evaluating 3D medical image segmentation: analysis, selection, and tool},
  author={Taha, Abdel Aziz and Hanbury, Allan},
  journal={BMC Medical Imaging},
  volume={15},
  pages={1--28},
  year={2015},
  publisher={Springer}
}

@inproceedings{zhou2019dup,
  title={Dup-net: Denoiser and upsampler network for 3d adversarial point clouds defense},
  author={Zhou, Hang and Chen, Kejiang and Zhang, Weiming and Fang, Han and Zhou, Wenbo and Yu, Nenghai},
  booktitle={Proceedings of the IEEE/CVF international conference on computer vision},
  pages={1961--1970},
  year={2019}
}

@inproceedings{carlini2017towards,
  title={Towards evaluating the robustness of neural networks},
  author={Carlini, Nicholas and Wagner, David},
  booktitle={2017 IEEE Symposium on Security and Privacy},
  pages={39--57},
  year={2017},
  organization={IEEE}
}

@inproceedings{tang2024symattack,
  title={Symattack: symmetry-aware imperceptible adversarial attacks on 3D point clouds},
  author={Tang, Keke and Wang, Zhensu and Peng, Weilong and Huang, Lujie and Wang, Le and Zhu, Peican and Wang, Wenping and Tian, Zhihong},
  booktitle={Proceedings of the 32nd ACM International Conference on Multimedia},
  pages={3131--3140},
  year={2024}
}

@inproceedings{tang2023deep,
  title={Deep manifold attack on point clouds via parameter plane stretching},
  author={Tang, Keke and Wu, Jianpeng and Peng, Weilong and Shi, Yawen and Song, Peng and Gu, Zhaoquan and Tian, Zhihong and Wang, Wenping},
  booktitle={Proceedings of the AAAI Conference on Artificial Intelligence},
  volume={37},
  pages={2420--2428},
  year={2023}
}

@inproceedings{zhou2018voxelnet,
  title={Voxelnet: End-to-end learning for point cloud based 3d object detection},
  author={Zhou, Yin and Tuzel, Oncel},
  booktitle={Proceedings of the IEEE Conference on Computer Vision and Pattern Recognition},
  pages={4490--4499},
  year={2018}
}

@inproceedings{zhu2017target,
  title={Target-driven visual navigation in indoor scenes using deep reinforcement learning},
  author={Zhu, Yuke and Mottaghi, Roozbeh and Kolve, Eric and Lim, Joseph J and Gupta, Abhinav and Fei-Fei, Li and Farhadi, Ali},
  booktitle={2017 IEEE International Conference on Robotics and Automation},
  pages={3357--3364},
  year={2017},
  organization={IEEE}
}

@inproceedings{landrieu2018large,
  title={Large-scale point cloud semantic segmentation with superpoint graphs},
  author={Landrieu, Loic and Simonovsky, Martin},
  booktitle={Proceedings of the IEEE Conference on Computer Vision and Pattern Recognition},
  pages={4558--4567},
  year={2018}
}

@article{sung2017complementme,
  title={ComplementMe: Weakly-supervised component suggestions for 3D modeling},
  author={Sung, Minhyuk and Su, Hao and Kim, Vladimir G and Chaudhuri, Siddhartha and Guibas, Leonidas},
  journal={ACM Transactions on Graphics},
  volume={36},
  pages={1--12},
  year={2017},
  publisher={ACM New York, NY, USA}
}

@inproceedings{shi2022shape,
  title={Shape prior guided attack: Sparser perturbations on 3d point clouds},
  author={Shi, Zhenbo and Chen, Zhi and Xu, Zhenbo and Yang, Wei and Yu, Zhidong and Huang, Liusheng},
  booktitle={Proceedings of the AAAI Conference on Artificial Intelligence},
  volume={36},
  pages={8277--8285},
  year={2022}
}

@article{guo2021pct,
  title={Pct: Point cloud transformer},
  author={Guo, Meng-Hao and Cai, Jun-Xiong and Liu, Zheng-Ning and Mu, Tai-Jiang and Martin, Ralph R and Hu, Shi-Min},
  journal={Computational Visual Media},
  volume={7},
  pages={187--199},
  year={2021},
  publisher={Springer}
}

@inproceedings{pang2025towards,
  title={Towards a 3D Transfer-based Black-box Attack via Critical Feature Guidance},
  author={Pang, Shuchao and Chen, Zhenghan and Zhang, Shen and Lu, Liming and Liang, Siyuan and Du, Anan and Zhou, Yongbin},
  booktitle={Proceedings of the IEEE/CVF International Conference on Computer Vision},
  pages={26912--26922},
  year={2025}
}

@inproceedings{uy2019revisiting,
  title={Revisiting point cloud classification: A new benchmark dataset and classification model on real-world data},
  author={Uy, Mikaela Angelina and Pham, Quang-Hieu and Hua, Binh-Son and Nguyen, Thanh and Yeung, Sai-Kit},
  booktitle={Proceedings of the IEEE/CVF International Conference on Computer Vision},
  pages={1588--1597},
  year={2019}
}

@inproceedings{li2022improving,
  title={Improving adversarial robustness of 3D point cloud classification models},
  author={Li, Guanlin and Xu, Guowen and Qiu, Han and He, Ruan and Li, Jiwei and Zhang, Tianwei},
  booktitle={European Conference on Computer Vision},
  pages={672--689},
  year={2022},
  organization={Springer}
}



\section*{Supplementary Material}

\setcounter{section}{0}
\renewcommand{\thefigure}{\thesection.\arabic{figure}}
\renewcommand{\thetable}{\thesection.\arabic{table}}
\renewcommand{\thesection}{\Alph{section}}

\makeatletter
\@addtoreset{figure}{section}
\@addtoreset{table}{section}
\makeatother

\section{Additional Ablation Studies on ModelNet40}
\begin{figure}[ht]
    \centering
    \includegraphics[width=0.95\linewidth]{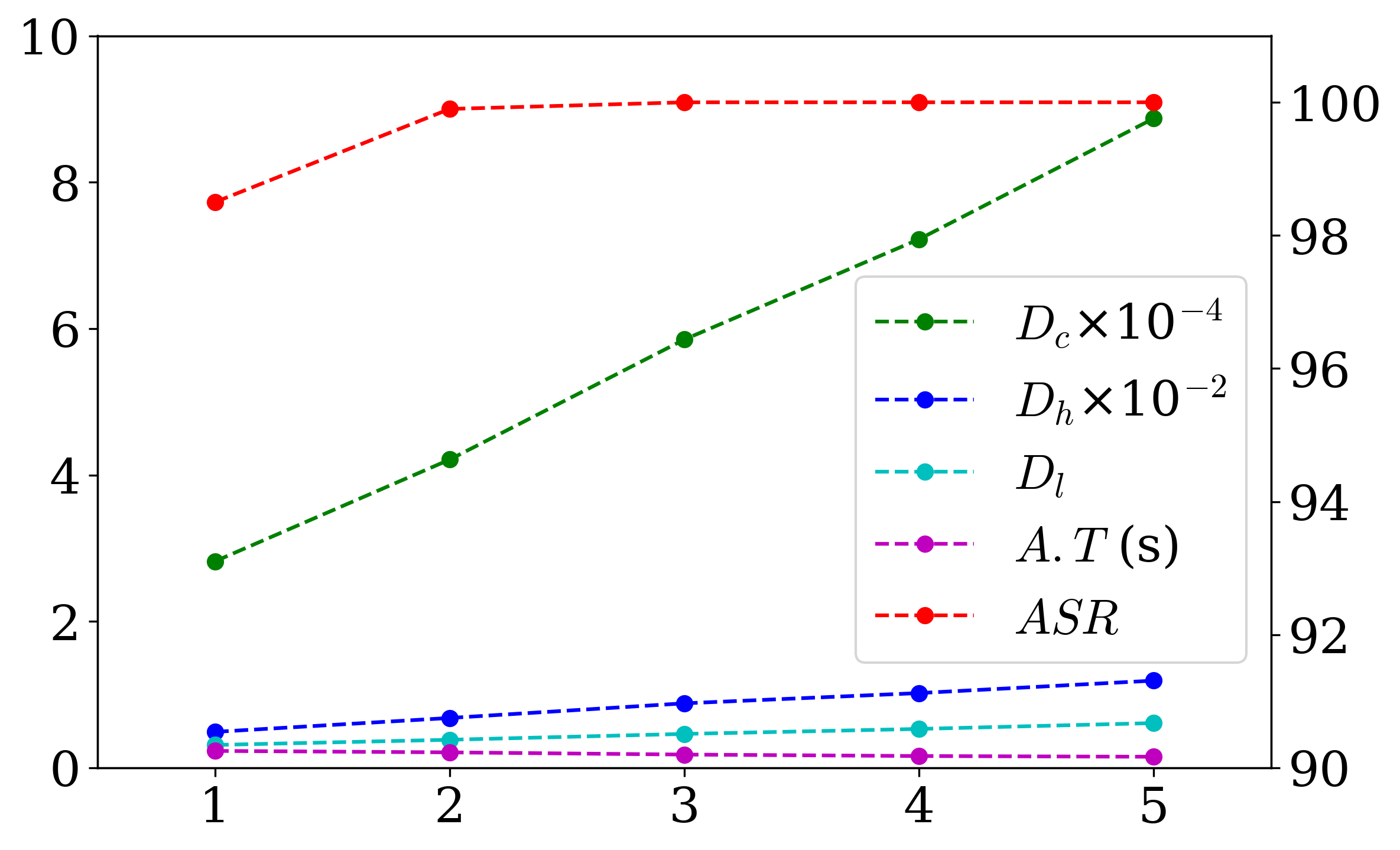}
    \vspace{-0.5em}
    \caption{Ablation study of the sensitivity coefficient $c$ on DGCNN~\cite{wang2019dynamic}. Left y-axis: $D_c~(10^{-4})$, $D_h~(10^{-2})$, $D_l$, $A.T~(s)$; right y-axis: ASR (\%).}
    \label{fig:supp_fig1_c}
    \vspace{-1em}
\end{figure}

\textbf{Effect of $c$. } As shown in Figure~\ref{fig:supp_fig1_c}, $c = 2$ stabilizes ASR while balancing imperceptibility ($D_c$, $D_h$, $D_l$) and efficiency (A.T). The larger $c$ slightly reduces A.T but significantly degrades stealthiness, offering negligible net benefit. Thus, we select $c = 2$.

\textbf{Weight Denominator Design.}
 As shown in Table~\ref{tab:supp_table1}, using $L_\infty$ in the weight denominator achieves the best trade-off: it produces the highest ASR (99.7\%), competitive imperceptibility ($D_c$, $D_h$, $D_l$), and the lowest A.T (0.24 s), making it the optimal choice.

\begin{figure}[ht]
    \centering
    \includegraphics[width=0.95\linewidth]{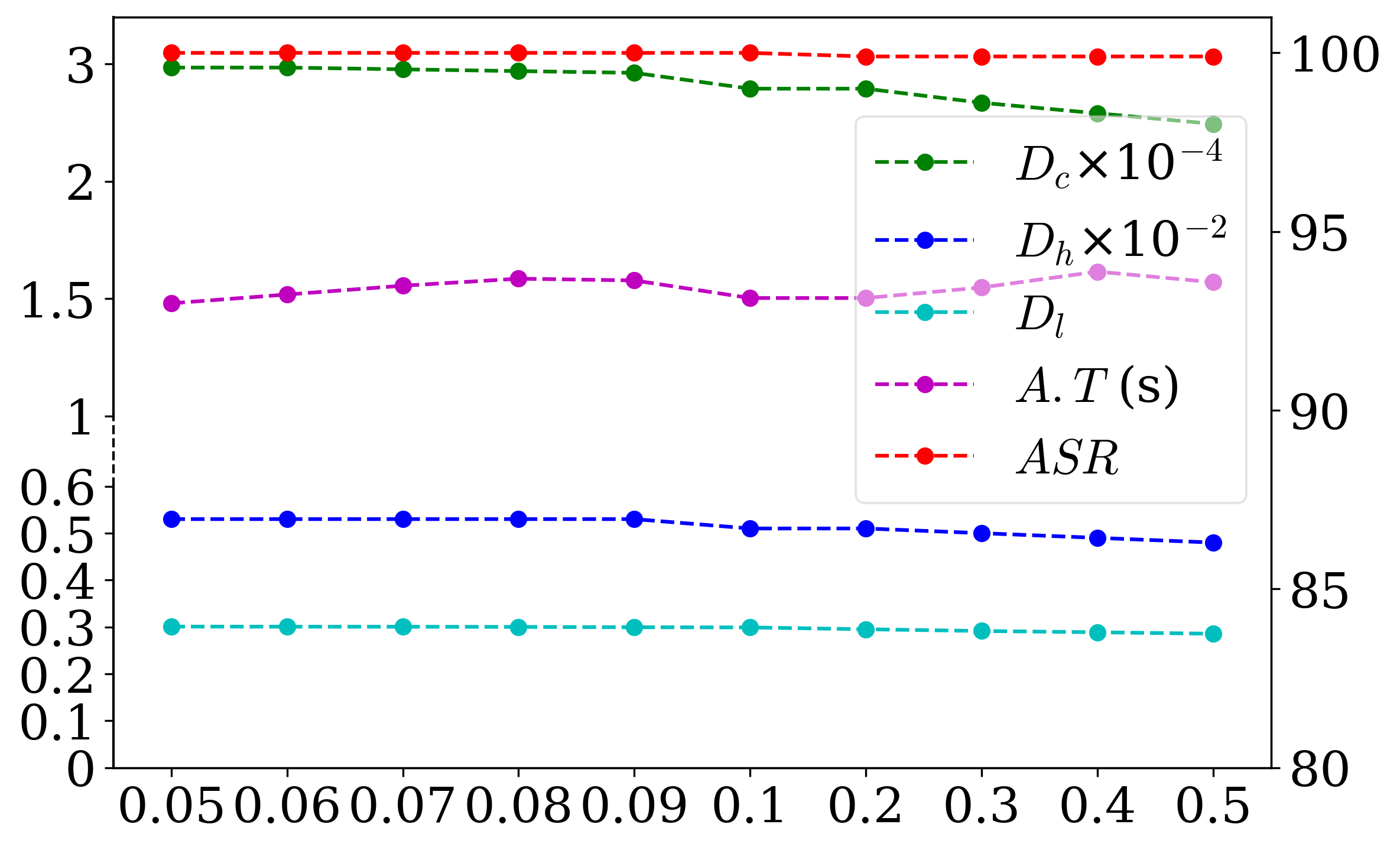}
    \vspace{-0.5em}
    \caption{Ablation study of $\lambda$ on DGCNN~\cite{wang2019dynamic}. Left y-axis: $D_c~(10^{-4})$, $D_h~(10^{-2})$, $D_l$, $A.T~(s)$; right y-axis: ASR (\%).}
    \label{fig:supp_fig2_lambda}
\end{figure}

\textbf{Effect of $\lambda$. } As shown in Figure~\ref{fig:supp_fig2_lambda}, on DGCNN~\cite{wang2019dynamic}, varying $\lambda$ in $[0.05, 0.5]$ produces stable ASR and imperceptibility ($D_c$, $D_h$, $D_l$), with only minor fluctuations in A.T. This indicates that $\lambda$ is insensitive to performance and requires only a reasonable setting. Thus, we adopt $\lambda = 0.1$ by default.

\begin{figure}[ht]
    \centering
    \includegraphics[width=0.95\linewidth]{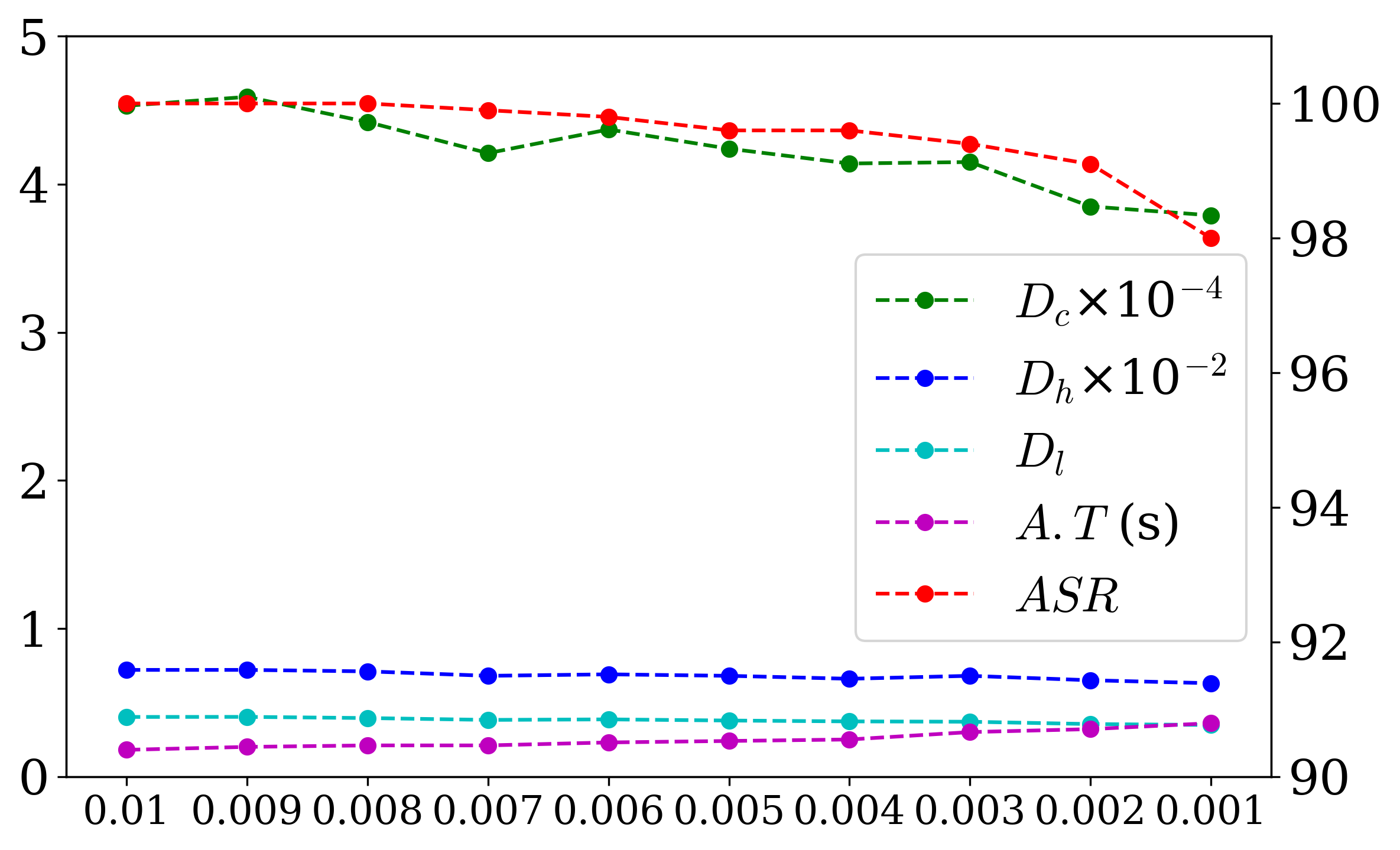}
    \vspace{-0.5em}
    \caption{Ablation study of the initial step size $\eta$ on DGCNN~\cite{wang2019dynamic}. Left y-axis: $D_c~(10^{-4})$, $D_h~(10^{-2})$, $D_l$, $A.T~(s)$; right y-axis: ASR (\%).}
    \label{fig:supp_fig3_initial_step}
\end{figure}

\textbf{Effect of the size of the initial step $\eta$.} As shown in Figure ~\ref{fig:supp_fig3_initial_step}, the size of the initial step $\eta$ has a pronounced impact: a larger $\eta$ increases both ASR and geometric distortion ($D_c$), while a smaller $\eta$ leads to significantly longer attack time (A.T). To strike an optimal balance among attack strength, imperceptibility, and efficiency, we set $\eta = 0.007$.

\begin{table}[htbp]
\centering
\caption{Ablation of weight denominator on DGCNN~\cite{wang2019dynamic}.}
\begin{tblr}{
  colspec = {cccccc},
  cells = {c, valign=m}, 
  hlines,
  vline{2} = {-}{},
}
Weight Denominator & {ASR\\(\%)} & {$D_c$\\($10\textsuperscript{-4}$)} & {$D_h$\\($10\textsuperscript{-2}$)} & $D_l$ & {A.T.\\(s)} \\
{$L_1$\\$L_2$\\$L_\infty$} & {64.1\\93.3\\99.7} & {2.03\\2.48\\3.20} & {0.40\\0.47\\0.55} & {0.2402\\0.2792\\0.3221} & {0.95\\0.49\\0.24}
\end{tblr}
\label{tab:supp_table1}
\end{table}

\section{Robustness Against 3D Defenses}
We evaluate WAAttack and SubAttack against three mainstream 3D defenses: SOR~\cite{zhou2019dup}, DUP-Net~\cite{zhou2019dup}, and CCN~\cite{li2022improving}, on ModelNet40 (MN)~\cite{wu20153d} and ScanObjectNN (SONN)~\cite{uy2019revisiting}. As shown in Table~\ref{tab:supp_table2}, our methods achieve state-of-the-art or second-best attack success rates (ASR) under SOR and CCN, significantly outperforming most baselines. Only under DUP-Net does performance slightly lag behind a few competitors, yet remains competitive. Notably, this strong performance holds across both synthetic (MN) and real-world scanned (SONN) data, highlighting the adaptability of our attacks to diverse input distributions. These results demonstrate the strong generalization and effectiveness of our approach across prevailing defense mechanisms.

\begin{table*}
\centering
\caption{Robustness evaluation of DGCNN~\cite{wang2019dynamic}  on ModelNet40 (MN)~\cite{wu20153d} and ScanObjectNN (SONN)~\cite{uy2019revisiting}. Best results are in \textbf{bold}, second-best are \underline{underlined}.}
\setlength{\tabcolsep}{2pt}
\begin{tabular}{c|c|cccccccccc} 
\hline
Dataset & Defense                                                      & 3D-Adv~\cite{xiang2019generating}                                                                   & JGBA~\cite{ma2020efficient}                                                                              & GeoA\textsuperscript{3}~\cite{wen2020geometry}                                                  & GSDA~\cite{hu2022exploring}                                                                     & SI-Adv~\cite{huang2022shape}                                                                   & PF-Attack~\cite{he2023generating}                                                      & CIM~\cite{zhang2024curvature}                                                                     & HiT-Adv~\cite{lou2024hide}                                                                 & WAAttack                                                         &SubAttack                                                                            \\ 
\hline
MN      & \begin{tabular}[c]{@{}c@{}}-\\SOR~\cite{zhou2019dup}\\DUP-Net~\cite{zhou2019dup}\\CCN~\cite{li2022improving}\end{tabular} & \begin{tabular}[c]{@{}c@{}}\textbf{100.0}\\89.2\\80.3\\91.7\end{tabular} & \begin{tabular}[c]{@{}c@{}}\textbf{100.0}\\\underline{97.9}\\89.5\\96.9\end{tabular}          & \begin{tabular}[c]{@{}c@{}}\textbf{100.0}\\97.4\\87.9\\95.3\end{tabular} & \begin{tabular}[c]{@{}c@{}}\textbf{100.0}\\97.2\\85.4\\96.8\end{tabular} & \begin{tabular}[c]{@{}c@{}}\textbf{100.0}\\96.4\\\underline{90.1}\\96.6\end{tabular} & \begin{tabular}[c]{@{}c@{}}\underline{99.7}\\95.1\\84.6\\94.9\end{tabular} & \begin{tabular}[c]{@{}c@{}}\underline{99.7}\\97.1\\89.1\\95.3\end{tabular}          & \begin{tabular}[c]{@{}c@{}}\textbf{100.0}\\94.3\\88.4\\91.9\end{tabular} & \begin{tabular}[c]{@{}c@{}}\underline{99.7}\\92.2\\\textbf{90.4}\\\underline{97.9}\end{tabular} & \begin{tabular}[c]{@{}c@{}}\textbf{100.0}\\\textbf{98.0}\\83.9\\\textbf{99.8}\end{tabular}  \\ 
\hline
SONN    & \begin{tabular}[c]{@{}c@{}}-\\SOR~\cite{zhou2019dup}\\DUP-Net~\cite{zhou2019dup}\\CCN~\cite{li2022improving}\end{tabular} & \begin{tabular}[c]{@{}c@{}}\textbf{100.0}\\90.4\\70.8\\85.9\end{tabular} & \begin{tabular}[c]{@{}c@{}}\textbf{100.0}\\98.1\\\textbf{85.7}\\97.3\end{tabular} & \begin{tabular}[c]{@{}c@{}}\textbf{100.0}\\94.1\\75.4\\97.7\end{tabular} & \begin{tabular}[c]{@{}c@{}}\textbf{100.0}\\94.5\\73.8\\98.4\end{tabular} & \begin{tabular}[c]{@{}c@{}}\underline{99.8}\\\underline{98.5}\\\underline{80.7}\\98.5\end{tabular}           & \begin{tabular}[c]{@{}c@{}}98.5\\96.8\\72.7\\96.7\end{tabular} & \begin{tabular}[c]{@{}c@{}}\underline{99.8}\\\textbf{99.1}\\79.1\\97.9\end{tabular} & \begin{tabular}[c]{@{}c@{}}98.7\\92.6\\81.8\\97.1\end{tabular}           & \begin{tabular}[c]{@{}c@{}}\underline{99.8}\\97.2\\79.3\\\underline{99.6}\end{tabular}          & \begin{tabular}[c]{@{}c@{}}\textbf{100.0}\\\textbf{99.1}\\73.3\\\textbf{99.7}\end{tabular}  \\
\hline
\end{tabular}
\label{tab:supp_table2}
\end{table*}

\section{Further Visualizations}
\begin{figure*}[t]
    \centering
    \includegraphics[width=1\linewidth]{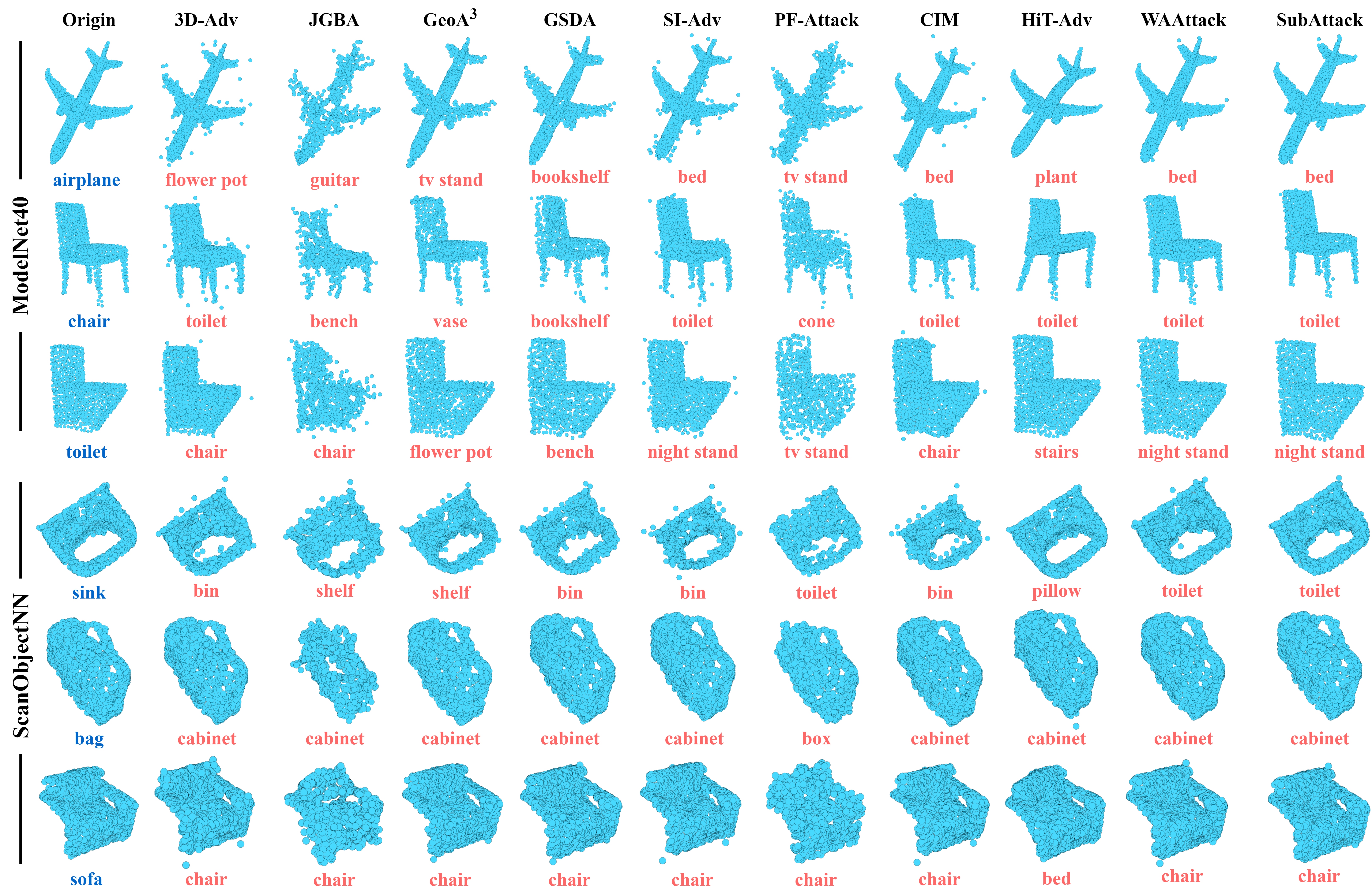}
    \caption{Visualization of adversarial point clouds generated by different attack methods on ModelNet40~\cite{wu20153d} and ScanObjectNN~\cite{uy2019revisiting} when targeting PointNet~\cite{qi2017pointnet}.}
    \label{fig:supp_visualization_pointnet}
\end{figure*}

This section presents visualizations of adversarial point clouds generated by various methods when attacking PointNet~\cite{qi2017pointnet}. As shown in Figure~\ref{fig:supp_visualization_pointnet}, consistent with the DGCNN~\cite{wang2019dynamic} results shown in the main paper, WAAttack and SubAttack produce perturbations that preserve geometric structure and exhibit no noticeable distortion or abnormal clustering, demonstrating strong imperceptibility. Notably, the magnitude of perturbations on ScanObjectNN~\cite{uy2019revisiting} is substantially smaller than on ModelNet40~\cite{wu20153d}, likely due to SONN’s real-world scanning noise and sparser geometry, which already provide a degree of natural camouflage for adversarial displacements. These visual comparisons further confirm the stability and superior robustness of our approach across different target architectures and data domains.

\end{document}